\documentclass{article}
\usepackage{arxiv}
\usepackage[utf8]{inputenc} 
\usepackage[T1]{fontenc}    
\usepackage{hyperref}       
\usepackage{url}            
\usepackage{booktabs}       
\usepackage{amsfonts}       
\usepackage{nicefrac}       
\usepackage{microtype}      
\usepackage{lipsum}		
\usepackage{graphicx}
\usepackage{natbib}
\usepackage{doi}
\usepackage{xcolor}         
\usepackage{times}
\usepackage{epsfig}
\usepackage{graphicx}
\usepackage{amsmath}
\usepackage{amssymb}
\usepackage{bbm}
\usepackage{microtype}
\usepackage{comment}
\usepackage{tabularray}
\usepackage{comment}
\usepackage{algorithm}
\usepackage{algpseudocode}
\usepackage{mathtools}
\usepackage{subcaption}
\def\E{{\mathbb{E}}}

\def\x{{\mathbf{x}}}
\def\y{{\mathbf{y}}}
\def\w{{\mathbf{w}}}
\def\z{{\mathbf{z}}}

\def\u{{\mathbf{u}}}
\def\v{{\mathbf{v}}}
\def\r{{\mathbf{r}}}
\def\s{{\mathbf{s}}}
\def\a{{\mathbf{a}}}
\def\g{{\mathbf{g}}}

\title{Probabilistic World Modeling with Asymmetric Distance Measure}

\author{ 
    {\hspace{1mm}Meng Song \thanks{Correspondence to \texttt{mes050@ucsd.edu}}}\\
	UC San Diego\\
}
\date{}

\begin{document}
\maketitle

\begin{abstract}
Representation learning is a fundamental task in machine learning, aiming at uncovering structures from data to facilitate subsequent tasks. However, what is a good representation for planning and reasoning in a stochastic world remains an open problem. In this work, we posit that learning a distance function is essential to allow planning and reasoning in the representation space. We show that a geometric abstraction of the probabilistic world dynamics can be embedded into the representation space through asymmetric contrastive learning. Unlike previous approaches that focus on learning mutual similarity or compatibility measures, we instead learn an asymmetric similarity function that reflects the state reachability and allows multi-way probabilistic inference. Moreover, by conditioning on a common reference state (e.g. the observer's current state), the learned representation space allows us to discover the geometrically salient states that only a handful of paths can lead through. These states can naturally serve as subgoals to break down long-horizon planning tasks. We evaluate our method in gridworld environments with various layouts and demonstrate its effectiveness in discovering the subgoals. 
\end{abstract}

\section{Introduction}
Learning good representations from the data plays a crucial role in the success of modern machine learning algorithms \cite{representation-learning}. It requires an AI system to have the ability to extract rich structures from the data and build a model of the world. What structures should be preserved, summarized, and what should be abstracted out is heavily dependent on the downstream tasks and learning objectives. For example, visual representation learning for recognition tasks mainly aims to expose the clustering structures in the representation space. Generative models such as VAEs \cite{vae}, GANs \cite{gan}, BERT \cite{bert}, diffusion models \cite{diffusion} and autoregressive models such as Transformers and RNNs \cite{transformer, rnn} capture the compressed information necessary for reconstructing the data from its corrupted version, or for directly generating the future predictions based on the data. Manifold learning methods such as LLE \cite{lle} and  ISOMAP \cite{ISOMAP} aim to learn a low-dimensional representation space preserving the local or global geometric structure of the data manifold. Interchangeably using one for another would not yield the best performance. In this paper, we consider the problem of \textit{what is a good representation for planning and reasoning in a stochastic world}. We will derive the problem formulation by delving into the definitions of planning and reasoning, along with the stochastic nature of the world.

Planning and reasoning typically involve an optimization process finding the most probable future outcomes, the most reasonable answers, or the most effective path to the goal. This optimization ability sets it apart from retrieving and interpolating good solutions seen in the training data, which imitation learning algorithms are good at. Instead, it enables problem-solving in new ways. \textit{To allow for such optimization, a fundamental requirement for the representation space is to have a distance function that accurately measures the probability of an event occurring given another, or the distance from the current state to the goal state.} For example, the optimal value function $V^*(\s, \g)$ in goal-conditioned RL, representing the optimal expected return when an agent is starting from state $\s$ and aiming to reach a particular goal state $\g$, can serve as a distance measure \cite{sorb}. However, a substantial body of previous work \cite{model-based-rl, video-as-language} has focused on learning the dynamics through a generative model that directly predicts the next state. Generating a predicted state on a detailed level is computationally expensive, especially when dealing with high-dimensional observations, and is more necessary for simulation than for planning. Furthermore, it is unclear whether the representation space learned by such reconstruction-based methods has a distance measure that adequately supports the optimization process in planning. 

The world inherently operates as a stochastic dynamical system, transiting from one state to another, often formulated as a stochastic process such as a Markov chain. A Markov chain induces a directed transition graph whose edges represent one-step transition probabilities from one state to another. By leveraging this transition graph, we can estimate the probability of transitioning from state $\x$ to state $\y$ within $C$ time steps, which we refer to as \textit{$C$-step reachability from $\x$ to $\y$}. Instead of looking at the single-step transition between neighbor states, $C$-step reachability considers all possible paths within $C$ steps from $\x$ to $\y$. A path on the transition graph can hold various interpretations in different contexts. For example, it can represent a sequence of events leading from event $\x$ to event $\y$. It can represent a reasoning chain starting from evidence or assumption $\x$ to conclusion $\y$. It can also represent a sequence of problem-solving steps beginning with problem $\x$ and resulting in the final solution $\y$. Therefore, $C$-step reachability allows us to do multi-way probabilistic inference and answer questions such as ``Given that event $\x$ has already occurred, how probable is it for event $\y$ to occur in the future, considering all possible ways the future could unfold?'', ``How likely to reach $\y$ from $\x$ considering all possible paths?'' 

By integrating the above thoughts, we formulate the representation learning problem for planning and reasoning in a stochastic world as the task of learning an embedding space. This space's distance function reflects the state reachability induced by a Markov chain. To address this challenge, we first establish a formal connection between the one-step transition matrix and the $C$-step reachability, then encode $C$-step reachability into an asymmetric similarity function by binary NCE \cite{nce-original, nce-ranking-binary}. A significant volume of prior studies on representation learning in NLP \cite{word2vec}, computer vision \cite{simcrl, moco, image-patch-embedding} and RL \cite{actionable, laplacian_rl} have implicitly modeled the co-occurrence statistics on an undirected graph and embed it by a single mapping function. In contrast to these prior works, we use a pair of embedding mappings to model the dual roles of a state in a directed graph: as an outgoing vertex and an incoming vertex. This approach assures an asymmetric similarity function reflecting the asymmetric even irreversible transition probabilities (e.g. the transition of food from raw to cooked), which is crucial for planning and event modeling tasks. 

Furthermore, the learned representation space also provides a geometric abstraction of the underlying directed graph in a perspective-dependent way. We find that by conditioning on a common reference state, a symmetric distance measure can be recovered to measure the point density in the representation space. This naturally gives rise to the notion of subgoals, which denote the geometrically salient states that only a handful of paths can lead through. The directional nature of the underlying transition graph reveals that the subgoal is inherently a relative concept and subject to change, which aligns well with our everyday experience but has not been explored by the previous works \cite{infobot, data-efficient-hrl}. For example, at a theme park entrance, a ticket is required for entry, but not for exit. Thus, the entrance serves as a subgoal upon arrival but not upon leaving. After formally defining the reference state conditioned distance measure, we can identify subgoal states as low-density regions using any density-based clustering algorithms. We demonstrate the effectiveness of our approach on solving the subgoal discovery task in a variety of gridworld environments. 

\section{Preliminaries}
\subsection{Markov chain and the directed transition graph}
A Markov chain on state space $\mathcal{S}$ can be thought of as a stochastic process traversing a directed graph where we start from vertex $\s_0 \sim \rho(S_0)$ and repeatedly follow an outgoing edge $\s_t \rightarrow \s_{t+1}$ with some probabilities. We denote the transition probability distribution of the Markov chain as $P(S_{t+1}|S_t)$ $(t=0,1,\dots)$. Thus, a Markov chain induces a weighted directed graph $G=(V, E, P)$ which is called the \textit{transition graph}. $V=\{\s \in \mathcal{S}\}$ is the vertex set, and $E=\{\s \rightarrow \s' \mid \s, \s' \in \mathcal{S}\}$ is a set of directed edges where each edge $\s \rightarrow \s'$ has probability $P(S_{t+1}=\s'|S_t=\s)$ as its weight. In this work, we consider the induced transition graph in the most general setting where loops are allowed, and a directed edge does not have to be paired with an inverse edge. 

\subsection{MDP and the environment graph} 
A Markov decision process (MDP) $\mathcal{M} = (\mathcal{S}, \mathcal{A}, P, r, \gamma, \rho)$ is an extension of a Markov chain with the addition of actions and rewards. The induced transition graph $G$ still has the states $\s \in \mathcal{S}$ as its vertex set, but the transition probability $P(S_{t+1}=\s'|S_t=\s)$ from vertex $\s$ to its neighbor $\s'$ now involves a two-stage transition process: we move from $\s$ to $\s'$ through some actions $\a$ according to both the environment dynamics $P(S_{t+1}|S_t, A_t)$ and the agent's policy $\pi(A_t|S_t)$:
\begin{equation} \label{eq:state_transition_general}
    P^{\pi}(S_{t+1}=\s'|S_t=\s)= \int P(S_{t+1}=\s' |S_t=\s, A_t=\a) \pi(A_t=\a | S_t=\s) d \a
\end{equation}
where the policy probability $\pi(A_t=\a | S_t=\s)$ acts as a weight of the environment transition probability $P(S_{t+1}=\s' |S_t=\s, A_t=\a)$.

In particular, when the policy is a uniform distribution for any state $\s$, \eqref{eq:state_transition_general} is irrelevant to the policy $\pi$, i.e.
\begin{equation} \label{eq:state_transition_no_policy}
    P(S_{t+1}=\s'|S_t=\s)= \frac{1}{|\mathcal{A}|}\int P(S_{t+1}=\s' |S_t=\s, A_t=\a) d \a
\end{equation}

We term this policy-agnostic transition graph as the \textit{environment graph} since its edge weights encode the environment dynamics unbiasedly and the graph can be fully induced from an MDP. In addition, we ignore the rewards as they are task-specific and do not necessarily reflect the structure of the environment.

\section{Problem formulation}
Suppose that a Markov chain $\mathcal{M}$ on state space $\mathcal{S}$ has an initial distribution $\rho(S_0)$ and an {\bf unknown} transition probability distribution $P(S_{t+1}|S_t)$. $G$ is the transition graph induced from $\mathcal{M}$. Given a set of $T$-step sequences drawn from $M$, i.e. $\mathcal{T}=\{(\s_0^i, \s_1^i, \ldots, \s_T^i)\}_{i=1}^N \sim \mathcal{M}$, our goal is to learn state representations whose distance function reflect the asymmetric vertex reachability on the underlying $G$.

In many cases, the state space $\mathcal{S}$ is either unknown or uncountable. As a result, enumerating the states at the very beginning of a Markov chain is infeasible. Therefore, we do not require a uniform initial distribution like the previous works \cite{node2vec, actionable, laplacian_rl}. On the contrary, we study the typical cases where the initial distribution $\rho(S_0)$ is highly concentrated, either a narrow Gaussian or a $\delta$ distribution centered at a specific state $\s_0$. 

\subsection{Vertex reachability}
Let random variables $U$ and $W$ represent any vertex on a directed graph $G=(V, E, P)$. The reachability from vertex $\u$ to vertex $\w$ can be defined as
\begin{equation} \label{eq:reachability_infinite}
    P(W=\w|U=\u) = \lim_{T \rightarrow \infty}\frac{1}{T}\sum_{t=1}^{T} P_t(S_t=\w|S_0=\u)
\end{equation}
\begin{equation} \label{eq:t-step-walk}
\begin{aligned}
P_t(S_t=\w|S_0=\u) &= \sum_{S_1, \dots, S_{t-1}} P(S_1, \ldots, S_{t-1}, S_t=\w | S_0=\u) \\
&= \sum_{S_1, \dots, S_{t-1}} \prod_{i=0}^{t-1} P(S_{i+1}|S_{i}) 
\end{aligned}
\end{equation}
where the edge $S_{i}\rightarrow S_{i+1}$ is in $E$ for all $0\leq i\leq t-1$. $P(S_{i+1}|S_{i})$ is the one-step transition probability distribution of $G$.  In other words, the sequence of vertices $S_0 \rightarrow S_1 \rightarrow \cdots \rightarrow S_t$ is a $t$-step walk on $G$ starting from $\u$ to $\w$. $P_t(S_t=\w|S_0=\u)$ is the probability of traveling from $\u$ to $\w$ in exactly $t$ steps. The reachability from $\u$ to $\w$ can be understood as the likelihood of reaching $\w$ from $\u$ within any number of steps. The derivation is done by thinking of $G$ as a Bayesian network. By definition, the reachability from vertex $\u$ to $\w$ may differ from the reachability from vertex $\w$ to $\u$, i.e. $P(W=\w|U=\u) \neq P(W=\u|U=\w)$.

\subsection{C-step approximation} \label{sec:c-step}
Computing the vertex reachability is often intractable since it requires enumerating all possible paths over a near-infinite horizon. In practice, we approximate  \eqref{eq:reachability_infinite} by looking ahead $C$ steps in a Markov chain. Formally, suppose that we have a $T$-step Markov chain $M=\{S_0, S_1, \dots, S_{T}\}$ with transition probability distribution $P(S_{t+1}|S_t) \triangleq P$, $t=0,1,\dots, T-1$. We denote any preceding random variables in the chain as the random variable $Y=\{S_i \mid 0 \leq i \leq T-1\}$, and any subsequent random variables within $C$ steps from $Y$ as the random variable $X=\{S_j \mid i < j \leq \min(i+C, T)\}$. The joint distribution $P(X=\w, Y=\u)$ represents the occurrence frequency of {\bf the ordered pair} $(\u, \w)$ on all the possible paths within $C$ steps. The reachability from $\u$ to $\w$ can be approximated as
\begin{equation} \label{eq:reachability_c_step}
\begin{aligned}
    P(W=\w|U=\u) &\approx P(X=\w|Y=\u) \\
    &\approx \frac{1}{C} \sum_{t=1}^C (P^{t})_{\u\w}
\end{aligned}
\end{equation}
where $P^{t}$ denotes the $t$-step transition probability distribution which is the product of $t$ one-step transition matrices $P$, i.e. $P^t = \underbrace{P P \cdots P}_{t}$. $(P^{t})_{\u\w} = P(S_{t} = \w \mid S_0 = \u)$. $P(X|Y)$ is also a stochastic matrix where each row sums to one and represents the reachability distribution starting from a specific state. 

The first approximation results from the fact that we reduce the near-infinite look-ahead steps to $C$ steps. The second line holds when $0 < C \ll T$ (See Appendix \ref{appendix:c-step-proof} for the proof). In this case, $P(X|Y)$ represents the accumulated transition probabilities up to $C$ steps, which align with the definition of reachability in \eqref{eq:reachability_infinite}. This suggests that when $1 \leq C \ll T$ is satisfied, one should use a larger $C$ to achieve better approximation precision. 

\section{Asymmetric contrastive representation learning}
We now aim to learn representations whose distance function encodes the asymmetric state reachability distribution $P(X|Y)$. In this section, we show how this problem can be formulated and addressed within the framework of noise contrastive estimation (NCE) \cite{original_nce}. 

NCE is a family of powerful methods for solving the density estimation problem while avoiding computing the partition function. The core idea of NCE is to transform the density estimation problem into a contrastive learning problem between the target distribution and a negative distribution. Based on different objective functions, NCE methods fall into two categories: binary NCE which discriminates between two classes and ranking NCE which ranks the true labels above the negative ones for the positive inputs \cite{nce-ranking-binary}. Although the ranking NCE is broadly used in the recent resurgence of contrastive representation learning \cite{cpc, simcrl, curl, moco}, we adopt binary NCE in this paper since it establishes a direct relationship between the scoring function and probability ratio.   

Concretely, suppose that there is a data set of $N$ trajectories drawn from a $T$-step Markov chain $\mathcal{M}$, i.e. $\mathcal{T}=\{(\s_0^i, \s_1^i, \dots, \s_T^i)\}_{i=1}^N \sim \mathcal{M}$. We construct the positive data set for binary NCE as $D^{+}=\{(\x^{+}_i, \y_i) \sim P(X, Y) \}_{i=1}^{N^{+}}$, and negative data set as $D^{-}=\{\x^{-}_i \sim P_{n}(X) \}_{i=1}^{N^{-}}$. $P(X,Y)$ denotes the aforementioned joint distribution of preceding random variable $Y$ and subsequent random variable $X$. $P_n(X)$ denotes a specified negative distribution on state space $\mathcal{S}$. We use $K$ to denote the ratio of negative and positive samples, i.e. $N^{-}=KN^{+}$. 

This yields a binary classification setting where a classifier $P(C|X, Y=\y; \theta)$ is trained to discriminate between positive samples from the conditional distribution $P(X| Y=\y)$ and the negative samples from $P_{n}(X)$, $\forall \y$. We use the label $C=\{1,0\}$ to denote the positive and negative classes, respectively. The training objective of binary NCE is to correctly classify both positive samples $D^{+}$ and negative samples $D^{-}$ using logistic regression:
\begin{equation} \label{eq:binary_nce_obj}
    \begin{aligned}
    \max_\theta J_{BI\_NCE}(\theta) 
    &= \max_\theta \E_{\x^{+}, \y \sim P(X,Y)} \log P(C=1 | \x^{+}, \y; \theta) + K \E_{\y \sim P_Y(Y)} \E_{\x^{-} \sim P_n(X)} \log P(C=0 | \x^{-}, \y; \theta) \\ 
    &= \max_\theta  \E_{\x^{+}, \y \sim P(X,Y)} \log \sigma(f_\theta(\x^{+},\y)) + K \E_{\y \sim P_Y(Y)} \E_{\x^{-} \sim P_n (X)} \log (1-\sigma(f_\theta(\x^{-},\y))) \\
    \end{aligned}
\end{equation}
where $P_Y(Y)$ denotes the marginal distribution of $Y$. $\sigma(\cdot)$ denotes the logistic function.

When the classifier is Bayes-optimal, we have 
\begin{equation} \label{eq:binary_nce_ratio}
\begin{aligned} 
\exp(f(X, Y=\y)) = \frac{P(X|Y=\y)}{K P_n(X)}, \quad \forall \y
\end{aligned}
\end{equation}
where $f(X, Y=\y)$ is the scoring function. (Proof in Appendix \ref{appendix:bayes-optimal-proof})

\subsection{Asymmetric encoders}
One of our key observations is that the scoring function $f(X=\x, Y=\y)$ in the conditional case of binary NCE can be treated as a similarity function between the embeddings of $\x$ and $\y$. Given that the underlying graph of a Markov chain is \textit{directed}, the reachability from vertex $\x$ to vertex $\y$ often differs from the reachability from vertex $\y$ to vertex $\x$. That is, 
\begin{equation} \label{eq:asymetric_reachability}
\begin{aligned}
P(X=\x|Y=\y) \neq P(X=\y|Y=\x) 
\end{aligned}
\end{equation}

Therefore, having an asymmetric similarity function becomes essential to mirror the inherent asymmetry in the reachability probabilities. In other words, the function form of $f$ should allow
\begin{equation} \label{eq:asymetric_distance}
\begin{aligned}
f(X=\x, Y=\y) \neq f(X=\y, Y=\x)
\end{aligned}
\end{equation}

One effective approach to accomplish this is by utilizing two separate encoders. Concretely, we use an encoding function $\phi: \mathcal{S} \rightarrow \mathcal{Z}$ to map the preceding random variable $Y$ to the embedding space $Z$ and use another encoding function $\psi: \mathcal{S} \rightarrow \mathcal{Z}$ to map the subsequent random variable $X$ to the same embedding space.
\begin{equation} \label{eq:z_distance}
\begin{aligned}
f(X=\x, Y=\y) = d(\psi(\x), \phi(\y))
\end{aligned}
\end{equation}
$d(\cdot)$ could be an arbitrary similarity measure in space $Z$, e.g. cosine similarity or negative $l_2$ distance.

In this paper, we opt for cosine similarity as measure $d(\cdot)$ in space $Z$ because it is well-bounded and facilitates stable convergence.
\begin{equation} \label{eq:cosine_similarity}
\begin{aligned} 
f(X=\x, Y=\y) = \frac{\psi(\x)}{\|\psi(\x)\|_2} \cdot \frac{\phi(\y)}{\|\phi(\y)\|_2}
\end{aligned}
\end{equation}
In this setup, \eqref{eq:binary_nce_ratio} becomes
\begin{equation} \label{eq:binary_nce_cosine}
\begin{aligned} 
\exp(\frac{\psi(\x)}{\|\psi(\x)\|_2} \cdot \frac{\phi(\y)}{\|\phi(\y)\|_2}) = \frac{P(X=\x|Y=\y)}{K P_n(X=\x)}
\end{aligned}
\end{equation}

Employing two separate encoders offers two advantages over a single encoder. Firstly, it guarantees the asymmetry of $f(\cdot, \cdot)$ irrespective of the choice of similarity measure $d(\cdot, \cdot)$ in the embedding space. Secondly, it ensures the asymmetry in a general sense without imposing constraints on the relationship between $\phi(\cdot)$ and $\psi(\cdot)$. On the contrary, prior works \cite{cpc, curl, c-learning-plan} employ a single encoder $\psi(\cdot)$ alongside a linear transformation $A$ to model time-series data, which can be viewed as a specific case of our approach where $\phi(\cdot)=A\psi(\cdot)$.

With slight adjustments, our method can also \textit{treat the directed graph as an undirected graph}. In such cases, we modify the range of random variable $Y$ and $X$ to $Y=\{S_i \mid 0 \leq i \leq T\}$ and $X=\{S_j \mid \max(i-C, 0) \leq j \leq \min(i+C, T), j \neq i \}$. Additionally, we use a single encoder $\phi(\cdot)$ to encode both $X$ and $Y$. $d(\cdot)$ could be any symmetric similarity measure, e.g. cosine similarity. As a result, we have
\begin{equation} \label{eq:symetric_distance}
\begin{aligned}
f(X=\x, Y=\y) &= f(X=\y, Y=\x) \\
&= d(\phi(\x), \phi(\y)) \\
&= \log \frac{P(X=\x|Y=\y)+P(X=\y|Y=\x)}{2K P_n(X=\x)}
\end{aligned}
\end{equation}
We transform the directed graph into an undirected one by enforcing the similarity function $f(\cdot)$ to encode the average of $P(X=\x|Y=\y)$ and $P(X=\y|Y=\x)$. Note that the reachability from $\x$ to $\y$ and the reachability from $\y$ to $\x$ may not be identical, meaning $P(X=\x|Y=\y)$ may not be equal to $P(X=\y|Y=\x)$. 

In fact, word embedding methods such as word2vec \cite{word2vec} and graph embedding methods such as node2vec \cite{node2vec} have implicitly performed the operations in \eqref{eq:symetric_distance} through the Skip-gram model. Learning undirected representations is sufficient when the downstream tasks are clustering, classification, etc. These applications rely on mutual similarity or compatibility measures insensible to the directionality. However, incorporating transition direction into representation learning is crucial for planning, reasoning, and event modeling. This becomes especially important when the transition is irreversible due to temporal order or causalities. For example, consider two states: $\y$=young and $\x$=old, transiting from young to old is certain while the reverse is impossible. In probabilistic language, we can express it as $P(X=\x|Y=\y) = 1$ and $P(X=\y|Y=\x) = 0$. The expected embeddings of these two states should correctly reflect this difference in reachability. Ideally, the state `young' should be very close to the state `old' in one scenario and be infinitely far away in the other. Averaging these two probabilities into $0.5$ would erroneously pull them together to the same distance in both cases.

\subsection{The choice of the negative distribution}
Although the NCE framework holds for an arbitrary negative distribution $P_n(X)$, different choices of $P_n(X)$ affect the similarity function $f(X,Y)$ in a meaningful way.  We experimented with several choices of the negative distribution $P_n(X)$: $P_X(\cdot)$ - the marginal distribution of $X$, $P_Y(\cdot)$ the marginal distribution of $Y$, and $U(X)$ - the uniform distribution of $X$. Among these options, $P_X(\cdot)$ is the only choice that consistently performs well across various values of chain length $T$ and step size $C$. In contrast, $U(X)$ performs the worst in most cases.  

It is worth noting that when $P_n(X)=P_X(X)$, the distance function $f(X,Y)$ encodes the pointwise mutual information (PMI) of an ordered pair $(\y, \x)$ up to a constant offset.
\begin{equation} \label{eq:pmi}
\begin{aligned} 
f(X=\x, Y=\y) &= \log \frac{P(X=\x|Y=\y)}{K P_X(X=\x)} \\
&= \log \frac{P(X=\x|Y=\y)}{P_X(X=\x)}+\log \frac{1}{K}
\end{aligned}
\end{equation}
where $\frac{1}{P_X(X=\x)}$ can be viewed as a weight of the reachability $P(X=\x|Y=\y)$, representing the inverse of the overall frequency of visits to the state $\x$. In other words, distribution $P(X=\x|Y=\y)$ is adjusted according to the rarity of the arrival state $\x$.

\section{Reference state conditioned distance measure}
Unlike standard representation learning methods which map each input into a single embedding vector, our approach maps each state $\s$ into two embedding vectors $\phi(\s)$ and $\psi(\s)$ in space $Z$. These two representations of $\s$ correspond to the two roles of a vertex in a directed graph respectively: $\phi(\s)$ corresponds to the role of $\s$ being an outgoing vertex and $\psi(\s)$ corresponds to the role of $\s$ being an incoming vertex. From the perspective of edges, each pair of vertices $\u$ and $\v$ can have two directed paths between them, $\u \rightarrow \v$ and $\v \rightarrow \u$, which have similarities $d(\phi(\u), \psi(\v))$ and $d(\phi(\v), \psi(\u))$ in the embedding space respectively. Therefore, the trained encoder $\phi(\cdot)$ and $\psi(\cdot)$ alongside the function $d(\cdot, \cdot)$ can fully capture the structure of the underlying directed graph $G$ in C-step approximation.

After training $\phi(\cdot)$, $\psi(\cdot)$, we can use them to perform inference tasks on a given set of states $\chi$. For instance, suppose that we are currently at state $\s$ and want to know which state is most likely to occur in the future. We can then identify it by finding the state closest to $\s$ in the latent space, i.e. solving $\arg \max_{\x \in \chi} d(\phi(\s), \psi(\x))$. Furthermore, with a defined action space, we can construct a directed graph on $\chi$ by evaluating $d(\phi(\cdot), \psi(\cdot))$. Feeding $G$ to any search algorithm, such as Dijkstra's algorithm, enables us to plan a shortest path from an initial state $\s_0$ to a goal state $\s_g$ in the latent space.

Moreover, we can still recover symmetric distance measures from the asymmetric similarity function $d(\cdot, \cdot)$. This allows us to perform clustering in the latent space without worrying about the metric disagreement. The ambiguity in metrics arises because each pair of vertices in a directed graph has two directional distances. To resolve this ambiguity, there must be a criterion to select one of them instead of simply averaging. Based on this principle, we demonstrate that a metric consensus in a directed graph can be achieved by comparing each pair of states to a common reference state $\r$. Therefore, we term it as a \textit{reference state conditioned distance measure}.   Formally, given a reference state $\r$, we define the pairwise latent distance between $\u$ and $\v$ with respect to $\r$ as $d_{\r}(\u, \v)$:
\begin{itemize}
    \item If $d(\phi(\r), \psi(\u)) \geq d(\phi(\r), \psi(\v))$, 
    \begin{equation} \label{eq:d_r_1}
        \begin{aligned} 
        d_{\r}(\u, \v) = 1-d(\phi(\u), \psi(\v))
        \end{aligned}
    \end{equation}
    \item Otherwise,
    \begin{equation} \label{eq:d_r_2}
        \begin{aligned} 
        d_{\r}(\u, \v) = 1-d(\phi(\v), \psi(\u))
        \end{aligned}
    \end{equation}
\end{itemize}
In other words, for a pair of states $\u$ and $\v$, we always choose the one closer to $\r$ as the starting state and choose the other one as the ending state to evaluate their distance. The symmetry of $d_{\r}(\cdot, \cdot)$ follows by the definition, i.e. $d_{\r}(\u, \v)=d_{\r}(\v, \u)$, $\forall \v, \u$. The reference state $\r$ could be any state in the state space $\mathcal{S}$, resulting in $|\mathcal{S}|$ symmetric distance measures $d_{\r}(\cdot, \cdot)$ extracted from $d(\cdot, \cdot)$. Unlike its undirected graph counterpart, there is no single and universal distance measure in a directed graph; instead, the measure varies depending on the observer's perspective. This definition of a reference-dependent measure $d_{\r}(\cdot, \cdot)$ allows us to find the salient geometric structure of the representation space from changing perspectives, which is meaningful in many real-world applications. We will demonstrate its usefulness in decision-making scenarios in the next section.   

\section{Subgoal discovery}
Subgoal is a fundamental concept introduced to tackle planning or decision-making problems using the divide-and-conquer strategy. It refers to the intermediate objectives or states that an agent aims to achieve on the way to reaching the ultimate goal, and are often used to break down complex or long-horizon tasks into shorter, simpler, more manageable parts.

While lacking a unified mathematical formulation, previous works have proposed various definitions of subgoals, each emphasizing its different roles in dividing the overall task, which include: midway states between the starting and goal state that are reachable by the current policy \cite{c-planning, sorb}; common states shared by successful trajectories \cite{diverse_density, gofar}; functionally salient states where policy distributions significantly change \cite{actionable};  and decision states where the goal state is informative to the decisions \cite{infobot}.  

In this work, we propose to use the concept of subgoals to denote the key states that only a handful of actions can lead through. Intuitively, the number of paths passing through a subgoal would decrease significantly. Following the definition, these states should be intrinsic to the geometric characteristics of the transition graph, independent of specific tasks and goals. More importantly, since the transition graph is inherently directed, subgoals may change as the observer's perspective varies. Therefore, unlike the previous works, we consider the subgoals as relative and subject to change, rather than as fixed and absolute entities. More precisely, we identify subgoals as the states that cause a sharp decrease in pairwise reachability, as perceived from the perspective of the agent's current state.

By the proposed representation learning method, we can convert the reachability defined on the original state space into the point density in the embedding space. The reduction in reachability can subsequently be translated into the identification of low-density regions, which density-based clustering algorithms can readily solve. In practice, we perform DBSCAN \cite{dbscan} on the representations to group the closely packed points while identifying the outliers that lie alone in the low-density regions as subgoals.

\section{Experiments}
\begin{figure*} [t] 
\centering
    \begin{subfigure}{0.15\textwidth}
        \includegraphics[width=\textwidth]{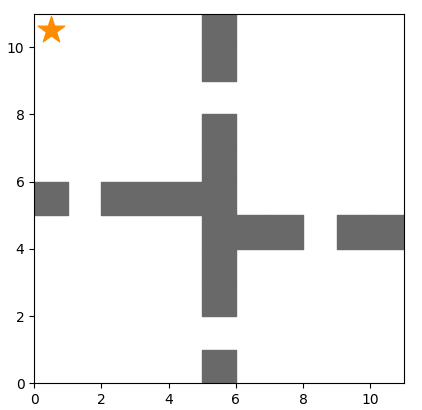}
        \caption{Four rooms}
    \end{subfigure}
    \begin{subfigure}{0.25\textwidth}
        \includegraphics[width=\textwidth]{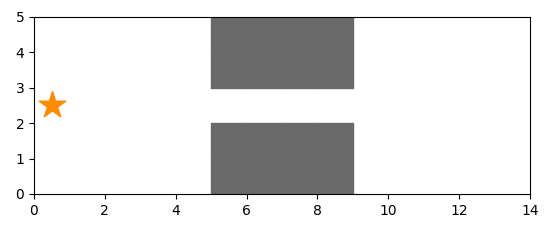}
        \caption{Dumbbell}
    \end{subfigure}
    \begin{subfigure}{0.2\textwidth}
        \includegraphics[width=\textwidth]{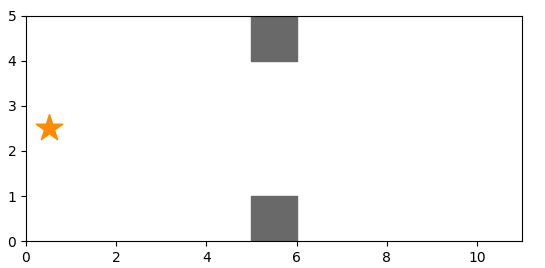}
        \caption{Wide door}
    \end{subfigure}
    \begin{subfigure}{0.17\textwidth}
        \includegraphics[width=\textwidth]{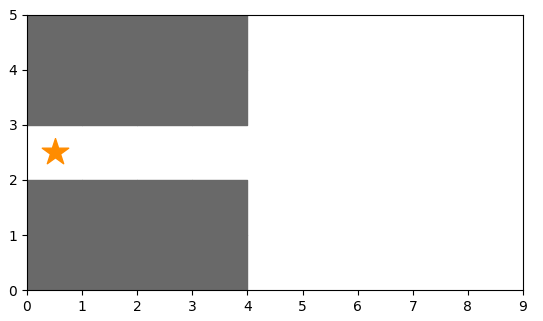}
        \caption{Flask}
    \end{subfigure}
    \begin{subfigure}{0.17\textwidth}
        \includegraphics[width=\textwidth]{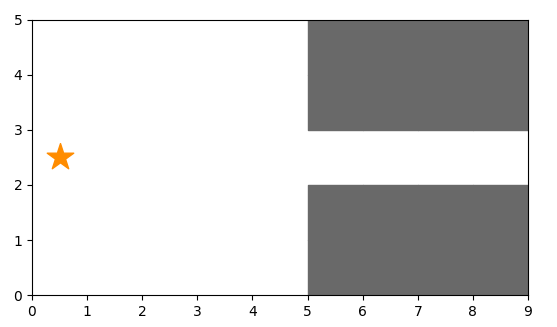}
        \caption{Nail}
    \end{subfigure}
    \caption{Gridworld environments: Grey areas indicate the walls. The yellow star indicates the initial state $\s_0$. When an agent collides with walls or attempts to move beyond the boundaries of the environment, its movement is blocked, and the agent remains in its current position.} \label{fig:envs}
\end{figure*}

We evaluate our representation learning method in five gridworld environments with various layouts: Four rooms, Dumbbell, Wide door, Flask, and Nail as shown in Figure \ref{fig:envs}. These environments are designed to encompass basic geometric configurations, serving as building blocks for constructing larger environmental graphs.  The states are 2D locations of the agent, $\s \in \mathbb{R}^2$. The action space includes five actions: left, right, up, down, and stop. In each environment, a data set of $N$ trajectories $\mathcal{T}=\{(\s_0^i, \s_1^i, \ldots, \s_T^i)\}_{i=1}^N$ is collected by a uniform policy starting from a fixed initial state $\s_0$. 

We train two encoders $\phi$ and $\psi$ on $\mathcal{T}$ according to the learning objective \eqref{eq:binary_nce_obj}. Both $\phi$ and $\psi$ are parameterized as a 2-layer MLP with latent space $\z \in \mathbb{R}^{64}$. We set the ratio of negative and positive samples $K=1$ throughout all the experiments. We found that  $K > 1$ performs worse because the classification becomes imbalanced. Moreover, $K=1$ also yields a precise correspondence between the learned asymmetric similarity function and PMI by removing the offset in \eqref{eq:pmi}. We use the marginal distribution of $X$ as the negative distribution for training, i.e. $P_n(X)=P_X(X)$, which we found performs the best. 

\subsection{t-SNE visualization of the learned representations}
\begin{figure*} 
\begin{subfigure}{\textwidth}
\centering
    \begin{subfigure}{0.2\textwidth}
        \includegraphics[width=\textwidth]{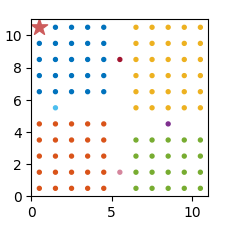}
    \end{subfigure}
    \begin{subfigure}{0.2\textwidth}
        \includegraphics[width=\textwidth]{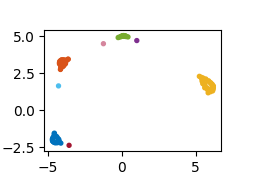}
    \end{subfigure}
    \hspace{10mm}
    \begin{subfigure}{0.2\textwidth}
        \includegraphics[width=\textwidth]{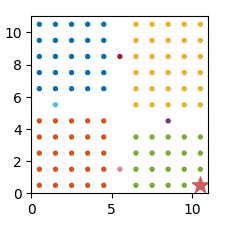}
    \end{subfigure}
    \begin{subfigure}{0.15\textwidth}
        \includegraphics[width=\textwidth]{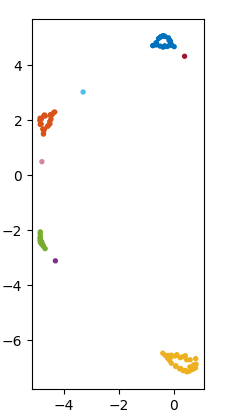}
    \end{subfigure}
    \caption{Four rooms}
\end{subfigure}
\begin{subfigure}{\textwidth}
\centering
    \begin{subfigure}{0.25\textwidth}
        \includegraphics[width=\textwidth]{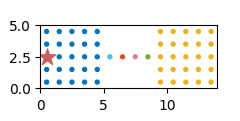}
    \end{subfigure}
    \begin{subfigure}{0.22\textwidth}
    \raisebox{2mm}{
        \includegraphics[width=\textwidth]{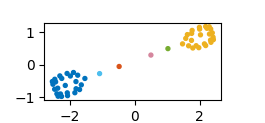}}
    \end{subfigure}
    \begin{subfigure}{0.25\textwidth}
        \includegraphics[width=\textwidth]{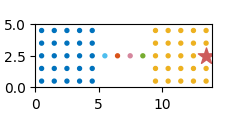}
    \end{subfigure}
    \begin{subfigure}{0.22\textwidth}
    \raisebox{3mm}{
        \includegraphics[width=\textwidth]{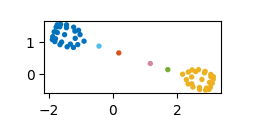}}
    \end{subfigure}
    \caption{Dumbbell}
\end{subfigure}
\begin{subfigure}{\textwidth}
\centering
    \begin{subfigure}{0.22\textwidth}
        \includegraphics[width=\textwidth]{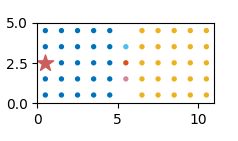}
    \end{subfigure}
    \begin{subfigure}{0.18\textwidth}
        \includegraphics[width=\textwidth]{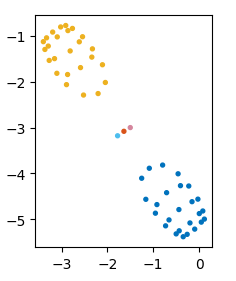}
    \end{subfigure}
    \hspace{5mm}
    \begin{subfigure}{0.22\textwidth}
        \includegraphics[width=\textwidth]{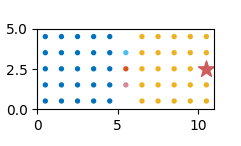}
    \end{subfigure}
    \begin{subfigure}{0.25\textwidth}
    \raisebox{5mm}{
        \includegraphics[width=\textwidth]{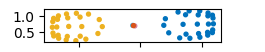}}
    \end{subfigure}
    \caption{Wide door}
\end{subfigure}
\begin{subfigure}{\textwidth}
\centering
    \begin{subfigure}{0.18\textwidth}
        \includegraphics[width=\textwidth]{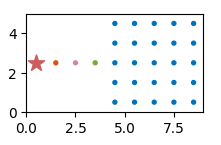}
    \end{subfigure}
    \begin{subfigure}{0.18\textwidth}
        \includegraphics[width=\textwidth]{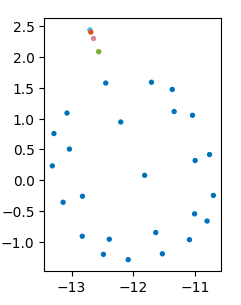}
    \end{subfigure}
    \hspace{15mm}
    \begin{subfigure}{0.18\textwidth}
        \includegraphics[width=\textwidth]{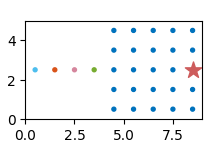}
    \end{subfigure}
    \begin{subfigure}{0.18\textwidth}
        \includegraphics[width=\textwidth]{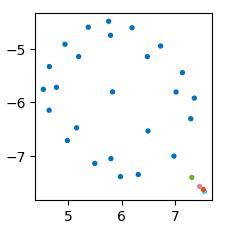}
    \end{subfigure}
    \caption{Flask}
\end{subfigure}
\begin{subfigure}{\textwidth}
\centering
    \begin{subfigure}{0.18\textwidth}
        \includegraphics[width=\textwidth]{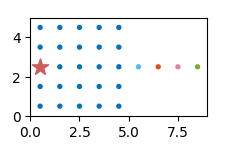}
    \end{subfigure}
    \begin{subfigure}{0.18\textwidth}
        \includegraphics[width=\textwidth]{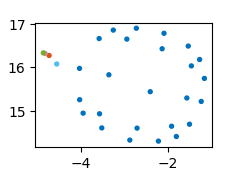}
    \end{subfigure}
    \hspace{15mm}
    \begin{subfigure}{0.18\textwidth}
        \includegraphics[width=\textwidth]{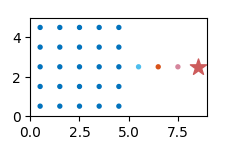}
    \end{subfigure}
    \begin{subfigure}{0.18\textwidth}
        \includegraphics[width=\textwidth]{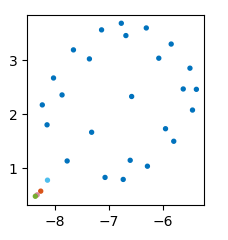}
    \end{subfigure}
    \caption{Nail}
\end{subfigure}
\caption{Visualization of the original states and the learned representations. The states are colored to visualize their position correspondences between two spaces. In each environment, we visualize the representation space from two different perspectives. The reference states $\r$ are indicated by the red stars. In each group, the left plot shows the original states in the 2D Euclidean space and the right plot shows the t-SNE projection of the learned representations. In all the experiments, we set approximation step size $C=16$, and train the encoders on a single episode of length $T=153600$.}\label{fig:visualization}
\end{figure*}

To examine the learned representations, we show the visualization of the learned representations and the original states in Figure \ref{fig:visualization}. We use t-SNE to project the 64-dimensional learned representations onto 2D plots based on the reference state conditioned distance measure $d_{\r}(\cdot, \cdot)$ defined in \eqref{eq:d_r_1} and \eqref{eq:d_r_2}. Across all the environments, the state space can be categorized into high-reachability regions such as rooms, and low-reachability regions such as long passages and various doorways (bottlenecks). Our learned representations can distinctly separate these regions according to their reachability with respect to the given reference state. 

\subsection{Subgoal discovery results}
\begin{figure*}
\begin{subfigure}{\textwidth}
\centering
    \begin{subfigure}{0.18\textwidth}
        \includegraphics[width=\textwidth]{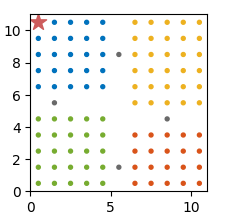}
    \end{subfigure}
    \begin{subfigure}{0.2\textwidth}
        \includegraphics[width=\textwidth]{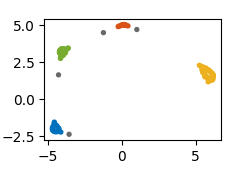}
    \end{subfigure}
    \hspace{10mm}
    \begin{subfigure}{0.18\textwidth}
        \includegraphics[width=\textwidth]{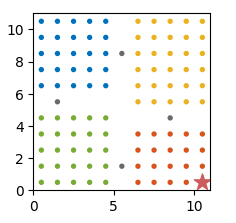}
    \end{subfigure}
    \begin{subfigure}{0.15\textwidth}
        \includegraphics[width=\textwidth]{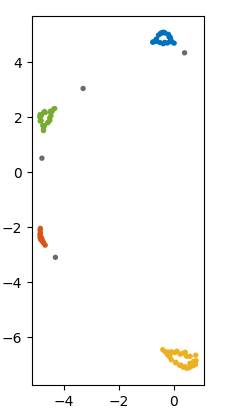}
    \end{subfigure}
    \caption{Four rooms}
\end{subfigure}
\begin{subfigure}{\textwidth}
\centering
    \begin{subfigure}{0.26\textwidth}
        \includegraphics[width=\textwidth]{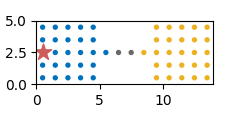}
    \end{subfigure}
    \begin{subfigure}{0.22\textwidth}
    \raisebox{3mm}{
        \includegraphics[width=\textwidth]{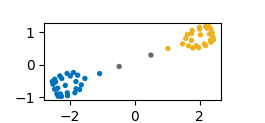}}
    \end{subfigure}
    \begin{subfigure}{0.25\textwidth}
        \includegraphics[width=\textwidth]{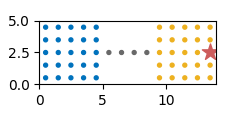}
    \end{subfigure}
    \begin{subfigure}{0.22\textwidth}
    \raisebox{3mm}{
        \includegraphics[width=\textwidth]{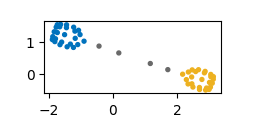}}
    \end{subfigure}
    \caption{Dumbbell}
\end{subfigure}
\begin{subfigure}{\textwidth}
\centering
    \begin{subfigure}{0.2\textwidth}
        \includegraphics[width=\textwidth]{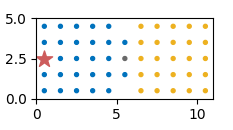}
    \end{subfigure}
    \begin{subfigure}{0.18\textwidth}
        \includegraphics[width=\textwidth]{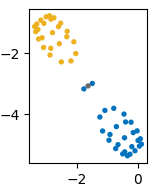}
    \end{subfigure}
    \hspace{5mm}
    \begin{subfigure}{0.2\textwidth}
        \includegraphics[width=\textwidth]{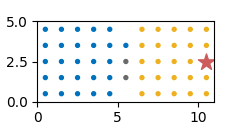}
    \end{subfigure}
    \begin{subfigure}{0.25\textwidth}
    \raisebox{5mm}{
        \includegraphics[width=\textwidth]{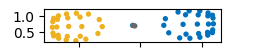}}
    \end{subfigure}
    \caption{Wide door}
\end{subfigure}
\begin{subfigure}{\textwidth}
\centering
    \begin{subfigure}{0.18\textwidth}
        \includegraphics[width=\textwidth]{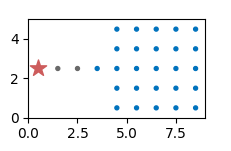}
    \end{subfigure}
    \begin{subfigure}{0.15\textwidth}
        \includegraphics[width=\textwidth]{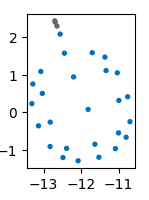}
    \end{subfigure}
    \hspace{15mm}
    \begin{subfigure}{0.18\textwidth}
        \includegraphics[width=\textwidth]{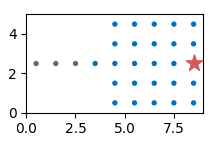}
    \end{subfigure}
    \begin{subfigure}{0.18\textwidth}
        \includegraphics[width=\textwidth]{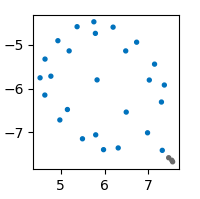}
    \end{subfigure}
    \caption{Flask}
\end{subfigure}
\begin{subfigure}{\textwidth}
\centering
    \begin{subfigure}{0.18\textwidth}
        \includegraphics[width=\textwidth]{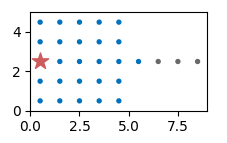}
    \end{subfigure}
    \begin{subfigure}{0.2\textwidth}
        \includegraphics[width=\textwidth]{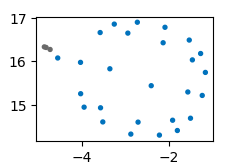}
    \end{subfigure}
    \hspace{15mm}
    \begin{subfigure}{0.18\textwidth}
        \includegraphics[width=\textwidth]{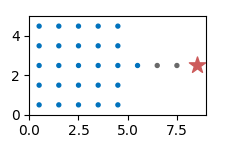}
    \end{subfigure}
    \begin{subfigure}{0.17\textwidth}
        \includegraphics[width=\textwidth]{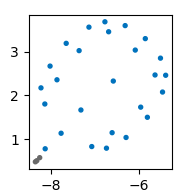}
    \end{subfigure}
    \caption{Nail}
\end{subfigure}
\caption{Subgoal discovery results. The states are colored according to the cluster labels in both the original space and the learned representation space. The gray states are subgoals. In each environment, we visualize the clustering results from two different perspectives. The reference states $\r$ are indicated by the red stars. In each group, the left plot shows the original states in the 2D Euclidean space and the right plot shows the t-SNE projection of the learned representations. In all the experiments, we set approximation step size $C=16$, and train the encoders on a single episode of length $T=153600$.} \label{fig:subgoal}
\end{figure*}

We now demonstrate that our learned distance measure $d_{\r}(\cdot, \cdot)$  enables easy identification of the subgoal states. We perform the DBSCAN clustering in the representation space according to $d_{\r}(\cdot, \cdot)$. The DBSCAN algorithm works by considering each point and expanding clusters from densely populated areas. During this procedure, the distance measure $d_{\r}(\cdot, \cdot)$ is evaluated to compute a point's local density and connectedness. Once cluster expansion is completed, the remaining points are labeled as noise which are just the subgoals we aim to identify. Figure \ref{fig:subgoal}
shows the subgoals and clusters found by our algorithm in each environment. The low-reachability regions such as doorways and passages have been successfully identified as subgoals and the rooms are decomposed into clusters.

\subsection{Ablation studies}
\subsubsection{The effect of approximation step size C}
\begin{figure*} 
\centering
    \begin{subfigure}{0.2\textwidth}
        \includegraphics[width=\textwidth]{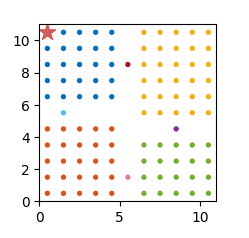}
        \caption{The original states}
    \end{subfigure}
    \begin{subfigure}{0.2\textwidth}
        \includegraphics[width=\textwidth]{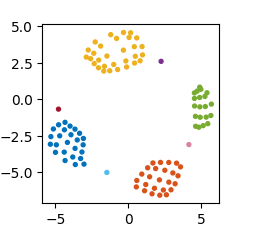}
        \caption{$C=1$}
    \end{subfigure}
    \begin{subfigure}{0.29\textwidth}
        \includegraphics[width=\textwidth]{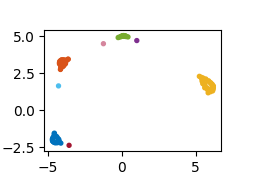}
        \caption{$C=16$}
    \end{subfigure}
    \begin{subfigure}{0.19\textwidth}
        \includegraphics[width=\textwidth]{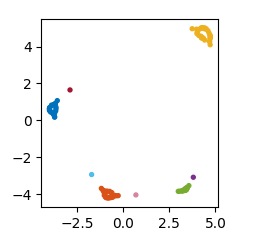}
        \caption{$C=64$}
    \end{subfigure}
    \caption{Visualization of the original states and the learned representations with different approximation step sizes $C$ in Four Rooms environment. The embeddings are projected to 2D plots by t-SNE. In all the experiments, we train the encoders on a single episode of length $T=153600$.} \label{fig:ablate_c}
\end{figure*}

To better understand the effect of step size $C$ in approximating the true reachability defined in \eqref{eq:reachability_infinite}, we compare the representations learned with $C=1,16,64$ in the four rooms environment. When $C=1$, the reachability is equivalent to the one-step transition probabilities. As the value of $C$ increases, the C-step reachability considers more and more possible paths between two states and thus progressively approaches the true reachability. As a result, the abstraction level of the representations becomes higher. This trend has been successfully captured in the learned distance measure $d_{\r}(\cdot, \cdot)$. As shown in Figure \ref{fig:ablate_c}, the rooms and doorways become further apart in the representation space as $C$ goes up. This is because the reachability contrast becomes sharper as longer paths are considered. These empirical observations align well with our theoretical derivations on the influence of step size $C$ in Section \ref{sec:c-step}. 

\subsubsection{The choice of negative distribution}
\begin{figure*}
\begin{subfigure}{\textwidth}
\centering
    \begin{subfigure}{0.25\textwidth}
    \raisebox{1mm}{
        \includegraphics[width=\textwidth]{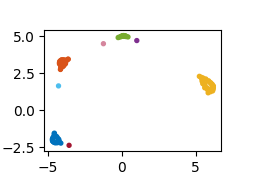}}
    \end{subfigure}
    \hspace{10mm}
    \begin{subfigure}{0.18\textwidth}
        \includegraphics[width=\textwidth]{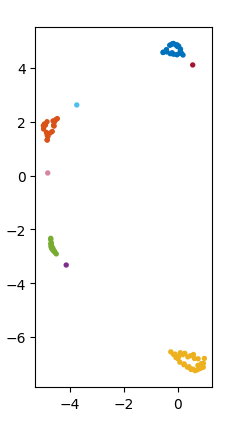}
    \end{subfigure}
    \hspace{15mm}
    \begin{subfigure}{0.2\textwidth}
    \raisebox{1mm}{
        \includegraphics[width=\textwidth]{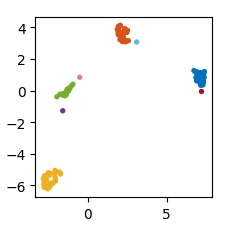}}
    \end{subfigure}
    \caption{$T=153600$}
\end{subfigure}

\begin{subfigure}{\textwidth}
\centering
    \begin{subfigure}{0.2\textwidth}
    \raisebox{1mm}{
        \includegraphics[width=\textwidth]{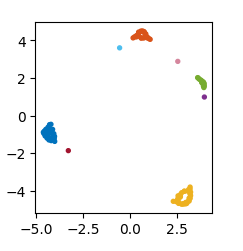}}
    \end{subfigure}
    \hspace{15mm}
    \begin{subfigure}{0.2\textwidth}
        \includegraphics[width=\textwidth]{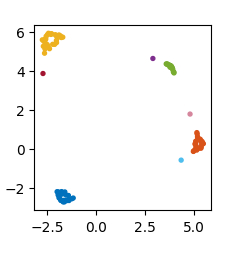}
    \end{subfigure}
    \hspace{15mm}
    \begin{subfigure}{0.22\textwidth}
    \raisebox{1mm}{
        \includegraphics[width=\textwidth]{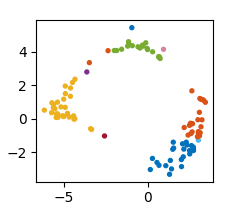}}
    \end{subfigure}
    \caption{$T=1024$}
\end{subfigure}

\begin{subfigure}{\textwidth}
\centering
    \begin{subfigure}{0.2\textwidth}
        \includegraphics[width=\textwidth]{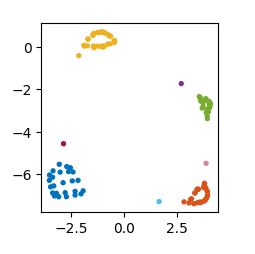}
    \end{subfigure}
    \hspace{15mm}
    \begin{subfigure}{0.21\textwidth}
        \includegraphics[width=\textwidth]{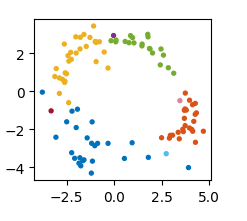}
    \end{subfigure}
    \hspace{15mm}
    \begin{subfigure}{0.21\textwidth}
    \raisebox{1mm}{
        \includegraphics[width=\textwidth]{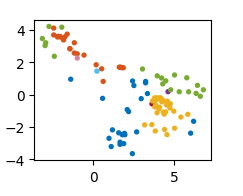}}
    \end{subfigure}
    \caption{$T=300$}
\end{subfigure}

\begin{subfigure}{\textwidth}
\centering
    \begin{subfigure}{0.2\textwidth}
        \includegraphics[width=\textwidth]{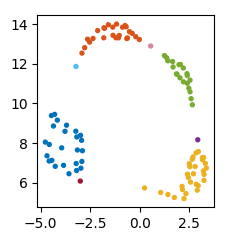}
    \end{subfigure}
    \hspace{15mm}
    \begin{subfigure}{0.21\textwidth}
    \raisebox{1mm}{
        \includegraphics[width=\textwidth]{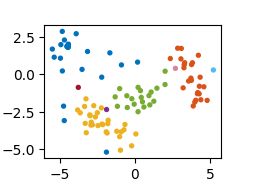}}
    \end{subfigure}
    \hspace{15mm}
    \begin{subfigure}{0.21\textwidth}
    \raisebox{1mm}{
        \includegraphics[width=\textwidth]{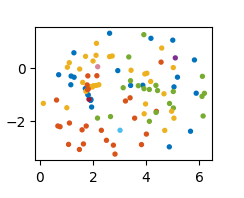}}
    \end{subfigure}
    \caption{$T=150$}
\end{subfigure}
\caption{Learned representations with different negative distributions in Four Rooms environment when $C=16$. Left column: $P_n(X)=P_X(X)$, Middle column: $P_n(X)=P_Y(X)$, Right column: $P_n(X)=U(X)$. Each row corresponds to the results with a different episode length.} \label{fig:ablate_pn_c=16}
\end{figure*}

\begin{figure*}
\begin{subfigure}{\textwidth}
\centering
    \begin{subfigure}{0.2\textwidth}
        \includegraphics[width=\textwidth]{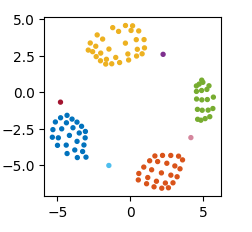}
    \end{subfigure}
    \hspace{15mm}
    \begin{subfigure}{0.2\textwidth}
        \includegraphics[width=\textwidth]{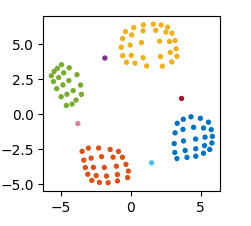}
    \end{subfigure}
    \hspace{15mm}
    \begin{subfigure}{0.2\textwidth}
        \includegraphics[width=\textwidth]{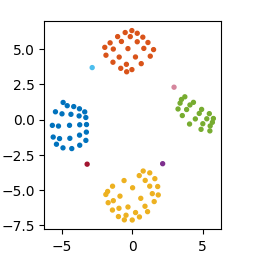}
    \end{subfigure}
    \caption{$T=153600$}
\end{subfigure}

\begin{subfigure}{\textwidth}
\centering
    \begin{subfigure}{0.2\textwidth}
        \includegraphics[width=\textwidth]{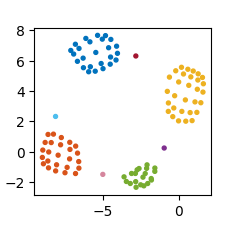}
    \end{subfigure}
    \hspace{15mm}
    \begin{subfigure}{0.2\textwidth}
        \includegraphics[width=\textwidth]{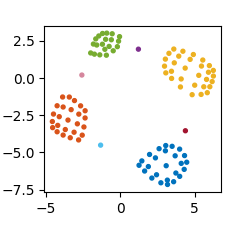}
    \end{subfigure}
    \hspace{15mm}
    \begin{subfigure}{0.22\textwidth}
        \includegraphics[width=\textwidth]{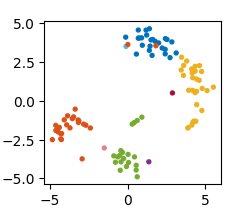}
    \end{subfigure}
    \caption{$T=1024$}
\end{subfigure}

\begin{subfigure}{\textwidth}
\centering
    \begin{subfigure}{0.2\textwidth}
        \includegraphics[width=\textwidth]{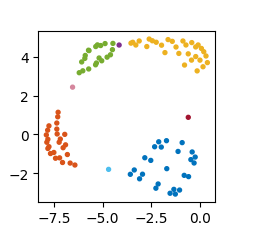}
    \end{subfigure}
    \hspace{15mm}
    \begin{subfigure}{0.21\textwidth}
        \includegraphics[width=\textwidth]{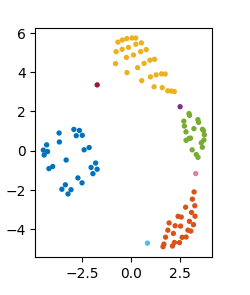}
    \end{subfigure}
    \hspace{15mm}
    \begin{subfigure}{0.2\textwidth}
        \includegraphics[width=\textwidth]{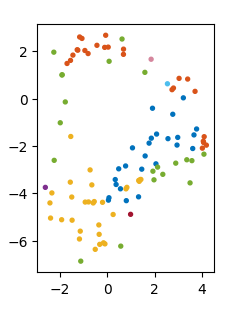}
    \end{subfigure}
    \caption{$T=300$}
\end{subfigure}

\begin{subfigure}{\textwidth}
\centering
    \begin{subfigure}{0.21\textwidth}
        \includegraphics[width=\textwidth]{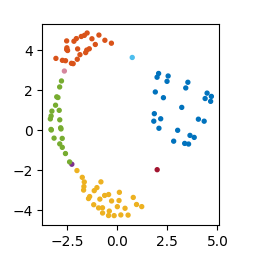}
    \end{subfigure}
    \hspace{15mm}
    \begin{subfigure}{0.21\textwidth}
    \raisebox{1mm}{
        \includegraphics[width=\textwidth]{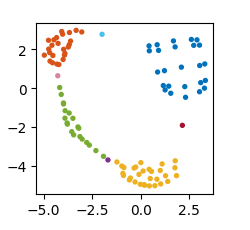}}
    \end{subfigure}
    \hspace{15mm}
    \begin{subfigure}{0.21\textwidth}
     \raisebox{1mm}{
        \includegraphics[width=\textwidth]{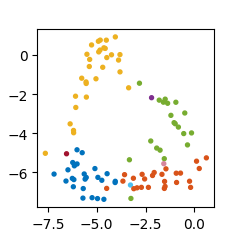}}
    \end{subfigure}
    \caption{$T=150$}
\end{subfigure}
\caption{Learned representations with different negative distributions in Four Rooms environment when $C=1$. Left column: $P_n(X)=P_X(X)$, Middle column: $P_n(X)=P_Y(X)$, Right column: $P_n(X)=U(X)$. Each row corresponds to the results with a different episode length.} \label{fig:ablate_pn_c=1}
\end{figure*}

We evaluate the representations learned with negative distributions $P_X(X)$, $P_Y(X)$ and $U(X)$ in the Four Rooms environment in Figure \ref{fig:ablate_pn_c=16} and \ref{fig:ablate_pn_c=1}. 
We observe that $P_X(X)$ consistently performs well across different values of episode length $T$ and step size $C$ while $U(X)$ performs the worst. We provide two hypotheses of why this phenomenon arises: (1) Using a negative distribution resembling the positive distribution helps train the classifier to reach Bayes optimum. $P_X(X)$ is more similar to $P(X|Y=\y)$ than $U(X)$. (2) The neural networks are secretly weighting the probability of $\x$ by its overall visitation frequencies when counting the occurrences of $\x$ given $\y$. The weight acts as an ``impression'' of a state. Therefore, choosing $P_X(X)$ as the denominator rather than $U(X)$ can counteract this weighting effect and recover the true $P(X|Y=\y)$.   

\section{Discussion and future work}
In this work, we study the problem of what is a good representation for planning and reasoning in a stochastic world and how to learn it. We discussed the importance of learning a distance measure in allowing planning and reasoning in the representation space. We modeled the world as a Markov chain and introduced the notion of $C$-step reachability on top of it to capture the geometric abstraction of the transition graph and allow multi-way probabilistic inference. We then showed how to embed the $C$-step abstraction of a Markov transition graph and encode the reachability into an asymmetric similarity function through conditional binary NCE. Based on this asymmetric similarity function, we developed a reference state conditioned distance measure, which enables the identification of geometrically salient states as subgoals. We demonstrated the quality of the learned representations and their effectiveness in subgoal discovery in the domain of gridworld. 

We leave the following topics for future work: (1) Extend the proposed method to learn the representations in continuous and high-dimensional state space. (2) Use the learned subgoals in hierarchical planning, reasoning, and HRL settings to solve long-horizon tasks. Unlike the model-free RL algorithms, our method effectively utilizes the data collected by random exploration to identify subgoals and build the environment model. Therefore, the sample efficiency is expected to be largely improved. (3) Use the learned similarity function as an intrinsic reward function to improve the performance of an RL agent. (4) Combine the proposed method of probabilistic directed graph embedding with the generative models to regularize the generated content and make them align well with how the world works.

\section{Appendix}
\subsection{Proof of C-step approximation} \label{appendix:c-step-proof}
Given a data set of $N$ trajectories drawn from a $T$-step Markov chain $M$, i.e. $\mathcal{T}=\{(\s_0^i, \s_1^i, \dots, \s_T^i)\}_{i=1}^N \sim \mathcal{M}$. Let $Y$ represent a preceding random variable in the chain $Y=\{S_i \mid 0 \leq i \leq T-1\}$ and $X$ represent a random variable subsequent to $Y$ within $C$ time steps. $X=\{S_j \mid i < j \leq \min(i+C, T)\}$. $P$ denotes the one-step transition matrix. We now derive $P(X|Y)$ in terms of $P$ as follows:

Based on the occurrences of the ordered pair $(\y, \x)$, we can derive the joint distribution of $X$ and $Y$ as 
\begin{equation} \label{eq:joint_count}
    \begin{aligned}
        P(X=\x, Y=\y) = \frac{\sum_{t=0}^{T-C}\sum_{i=t+1}^{t+c} n(S_i=\x, S_t=\y)+\sum_{t=T-C+1}^{T-1}\sum_{i=t+1}^{T}n(S_i=\x, S_t=\y)}{N(T-C+1)C + N\sum_{t=1}^{C-1}t }
    \end{aligned}
\end{equation}
where $n(S_i=\x, S_t=\y)$ denotes the number of times $S_i=\x$ and $S_t=\y$ occur in data set $\mathcal{T}$.

Similarly, by counting the occurrences of each specific state $\y$, the marginal distribution of $Y$ can be derived as
\begin{equation} 
    \begin{aligned} \label{eq:marginal_y_count}
        P_Y(Y=\y) = \frac{C\sum_{t=0}^{T-C}n(S_t=\y)+\sum_{t=T-C+1}^{T-1} (T-t) n(S_t=\y)}{N(T-C+1)C + N\sum_{t=1}^{C-1}t}
    \end{aligned}
\end{equation}
where $n(S_t=\y)$ denotes the number of times $S_t=\y$ occurs in data set $\mathcal{T}$.

By the definition of the Markov chain, we have
\begin{equation} \label{eq:count_redistribute}
    \begin{aligned}
    n(S_i=\x, S_t=\y) = n(S_t=\y) (P^{i-t})_{\y\x} \quad (i=t+1, t+2, \ldots)
    \end{aligned}
\end{equation}

Plugging \eqref{eq:count_redistribute} into \eqref{eq:joint_count}, we have
\begin{equation}
    \begin{aligned} \label{eq:joint_p}
        P(X=\x, Y=\y) = \frac{\sum_{t=0}^{T-C}n(S_t=\y) \sum_{i=1}^C (P^{i})_{\y\x} +\sum_{t=T-C+1}^{T-1}\sum_{i=t+1}^{T}n(S_t=\y)(P^{i-t})_{\y\x}}{N(T-C+1)C + N\sum_{t=1}^{C-1}t }
    \end{aligned}
\end{equation}

Given \eqref{eq:joint_p} and \eqref{eq:marginal_y_count}, we can derive the conditional distribution as
\begin{equation}
    \begin{aligned}
    P(X=\x \mid Y=\y) &= \frac{P(X=\x, Y=\y)}{P(Y=\y)} \\
    &= \frac{\sum_{t=0}^{T-C}n(S_t=\y) \sum_{i=1}^C (P^{i})_{\y\x} +\sum_{t=T-C+1}^{T-1}\sum_{i=t+1}^{T}n(S_t=\y)(P^{i-t})_{\y\x}}{C\sum_{t=0}^{T-C}n(S_t=\y)+\sum_{t=T-C+1}^{T-1} (T-t) n(S_t=\y)}
    \end{aligned}
\end{equation}

When $0 < C \ll T$, we can further drop the second term of the nominator and denominator, which is equivalent to changing the range of $Y$ to $Y=\{S_i \mid 0 \leq i \leq T-C\}$. That is,
\begin{equation}
    \begin{aligned}
    P(X=\x \mid Y=\y) &\approx \frac{\sum_{t=0}^{T-C}n(S_t=\y) \sum_{i=1}^C (P^{i})_{\y\x} }{C\sum_{t=0}^{T-C}n(S_t=\y)} \\
    &= \frac{1}{C} \sum_{t=1}^C (P^{t})_{\y\x}
    \end{aligned}
\end{equation}

\subsection{Proof of Bayes optimal scoring function} \label{appendix:bayes-optimal-proof}
Suppose that we are training a binary classifier $P(C|X, Y=\y; \theta)$ to discriminate between positive data distribution $P(X| Y=\y)$ and negative data distribution $P_{n}(X)$, $\forall \y$. The ratio of negative and positive samples is $K$. The Bayes optimal classifier $P(C|X, Y=\y; \theta)$ is a maximum a posteriori (MAP) hypothesis satisfying
\begin{equation} 
    \begin{aligned}
    \frac{P(X| Y=\y)}{P_{n}(X)} 
    &=
    \frac{P(X|Y=\y, C=1)}{P(X|C=0)} \\
    &=  \frac{P(C=1|X, Y=\y; \theta)}{P(C=0 | X, Y=\y; \theta)} \frac{P(C=0)}{P(C=1)} \\
    & = \frac{P(C=1|X, Y=\y; \theta)}{P(C=0 | X, Y=\y; \theta)} K \\
    & = \frac{P(C=1|X, Y=\y; \theta)}{1 - P(C=1|X, Y=\y; \theta)} K, \quad \forall \y
    \end{aligned}
\end{equation}

\begin{equation} \label{eq:binary_bayes_classifier}
    \begin{aligned}
    P(C=1|X, Y=\y; \theta) &= \frac{P(X| Y=\y)}{P(X| Y=\y)+K P_{n}(X)}, \quad \forall \y
    \end{aligned}
\end{equation}
When the classifier is a logistic function, we have
\begin{equation} \label{eq:binary_logistic_reg_pos}
    \begin{aligned}
    P(C=1|X, Y=\y; \theta) = \frac{1}{1+\exp(-f_\theta(X, Y=\y))}, \quad \forall \y
\end{aligned}
\end{equation}
Plugging \eqref{eq:binary_logistic_reg_pos} into \eqref{eq:binary_bayes_classifier}, we have
\begin{equation}
\begin{aligned} 
\exp(f_\theta(X, Y=\y)) = \frac{P(X|Y=\y)}{K P_n(X)}, \quad \forall \y
\end{aligned}
\end{equation}

\bibliographystyle{unsrtnat}
\bibliography{references}

\end{document}